\pdfoutput=1
\PassOptionsToPackage{table,xcdraw,dvipsnames}{xcolor}

\documentclass[11pt]{article}

\usepackage{acl}

\usepackage{times}
\usepackage{latexsym}

\usepackage[T1]{fontenc}

\usepackage[utf8]{inputenc}

\usepackage{microtype}

\usepackage{inconsolata}

\usepackage{rotating}
\usepackage[inline]{enumitem}
\usepackage{float}
\usepackage[fleqn]{amsmath}
\usepackage{algorithm}
\usepackage{algpseudocode}
\usepackage[inline]{enumitem}
\usepackage{graphicx}
\usepackage{tikz-cd}
\usepackage{multirow} 
\usepackage{cprotect}
\usepackage{comment}
\usepackage[T1]{fontenc}
\usepackage{caption}
\usepackage{diagbox}
\usepackage{listings}
\usepackage{wrapfig}
\usepackage{stfloats}
\usepackage{slashbox}
\definecolor{darkbrown}{rgb}{0.4, 0.26, 0.13}
\definecolor{burgundy}{rgb}{0.5, 0.0, 0.13}
\usepackage{balance}
\usepackage{amsfonts}
\usepackage{subcaption,graphicx}

\definecolor{light-gray}{gray}{0.95}
\newcommand{\codec}[1]{\colorbox{light-gray}{\texttt{#1}}}
\newcommand{\code}[1]{\texttt{#1}}

\title{EXPLORER: Exploration-guided Reasoning\\ for Textual Reinforcement Learning}

\author{ 
        Kinjal Basu,\textsuperscript{1}
        Keerthiram Murugesan,\textsuperscript{1}
        Subhajit Chaudhury,\textsuperscript{1} 
        Murray Campbell,\textsuperscript{1} \\
        {\bf Kartik Talamadupula,\textsuperscript{2} 
        and Tim Klinger,\textsuperscript{1} }\\ 
        \textsuperscript{1} IBM Research \hspace{10mm} \textsuperscript{2} Symbl.ai \\
        {\texttt{ \{kinjal.basu, keerthiram.murugesan, subhajit\}@ibm.com}},\\ {\texttt{ kartik.t@symbl.ai}}, {\texttt{ \{mcam, tklinger\}@us.ibm.com }}    
    }
\def\ex{EXPLORER }

\begin{document}
\maketitle
\begin{abstract}
Text-based games (TBGs) have emerged as an important collection of NLP tasks, requiring reinforcement learning (RL) agents to combine natural language understanding with reasoning. A key challenge for agents attempting to solve such tasks is to generalize across multiple games and demonstrate good performance on both seen and unseen objects. Purely deep-RL-based approaches may perform well on seen objects; however, they fail to showcase the same performance on unseen objects. Commonsense-infused deep-RL agents may work better on unseen data; unfortunately, their policies are often not interpretable or easily transferable. To tackle these issues, in this paper, we present EXPLORER\footnote{Code available at: \href{https://github.com/kinjalbasu/explorer}{https://github.com/kinjalbasu/explorer}} which is an exploration-guided reasoning agent for textual reinforcement learning. \ex is neuro-symbolic in nature, as it relies on a neural module for exploration and a symbolic module for exploitation. It can also learn generalized symbolic policies and perform well over unseen data. Our experiments show that \ex outperforms the baseline agents on Text-World cooking (TW-Cooking) and Text-World Commonsense (TWC) games. 
\end{abstract}

\section{Introduction}


 \begin{figure}[t]
    \includegraphics[scale = 0.92]{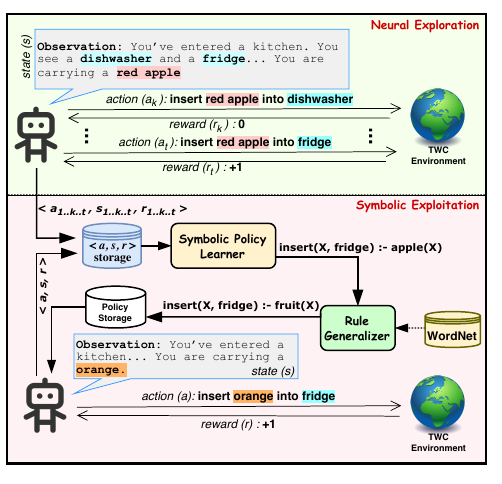}
    \caption{An overview of the EXPLORER agent's dataflow on a TWC game. In EXPLORER, the neural module is responsible for \textit{\textbf{exploration}} and collects <\textit{action}, \textit{state}, \textit{reward}> pairs, whereas the symbolic module learns the rules and does the \textbf{\textit{exploitation}} using commonsense knowledge from WordNet.}
    \label{fig:explorer_example}
\end{figure}

Natural language plays a crucial role in human intelligence and cognition. To study and evaluate the process of language-informed sequential decision-making in AI agents, text-based games (TBGs) have emerged as important simulation environments, where the states and actions are usually described in natural language. To solve game instances, an agent needs to master both natural language processing (NLP) and reinforcement learning (RL). At a high level, existing RL agents for TBGs can be classified into two classes: (a) rule-based agents, and (b) neural agents. Rule-based agents such as NAIL~\cite{nail} rely heavily on prior predefined knowledge. This makes them less flexible and adaptable. To overcome the challenges of rule-based agents, in recent years, with the advent of new deep learning techniques, significant progress has been made on neural agents~\cite{narasimhan2015language, adhikari2020learning}. However, these frameworks also suffer from a number of shortcomings. First, from deep learning, they inherit the need for very large training sets, which entails that they learn slowly. Second, they are brittle in the sense that a trained network may show good performance with the entities that are seen in the training instances, yet it performs very poorly in a very similar environment with unseen entities. Additionally, the policies learned by these neural RL agents are not interpretable (human-readable).

In this paper, we introduce \textbf{\ex} for TBGs that utilizes the positive aspects of both neural and symbolic agents. The \ex is based on two modules - neural and symbolic, where the neural module is mainly responsible for \textit{exploration} and the symbolic module does the \textit{exploitation}. An overview of the EXPLORER agent can be found in Figure \ref{fig:explorer_example}. 
A key advantage of \ex is that it has a scalable design that can integrate any neural module and can build the symbolic module upon it. 
For the symbolic module, instead of using predefined prior knowledge, \ex learns its symbolic policies by leveraging reward and action pairs while playing the game. These policies are represented using a declarative logic programming paradigm --- Answer Set Programming (ASP) \cite{lifschitz2019answer}, which allows the policies to be interpretable and explainable. Due to its non-monotonic nature and efficient knowledge representation ability, ASP has proven its efficiency in NLP research \cite{basu2020aqua, basu2021knowledge, pendharkar2022asp, zeng2024automated}; Commonsense reasoning research \cite{gupta2023building, kothawade2021auto}; and NLP + RL research \cite{sdrl, basu2022symbolic, asp-rl-sridhar, mitra-asp, peorl}. We believe non-monotonic reasoning (NMR) \cite{gelfond1988stable, nmr} is a crucial capability in partially observable worlds, as the agent's beliefs can change in the presence of new information and examples. Importantly, with the help of an exception learner (illustrated in Section \ref{sec:exc-learner}), \ex learns the symbolic policies as default theories so that the agent can perform NMR, and the policies remain consistent with the agent's findings.

After learning the symbolic policies, \ex can {\em lift} or variablize the rules using WordNet \cite{wn} to generalize them. By generalizing the symbolic policies, we find that \ex overcomes the challenge of getting poor performance over unseen entities or out-of-distribution (OOD) test sets, as the unseen objects are similar in nature to the training objects and occur under the same class in WordNet. 

\begin{figure}[t]
    \centering
    \includegraphics[scale = 0.45]{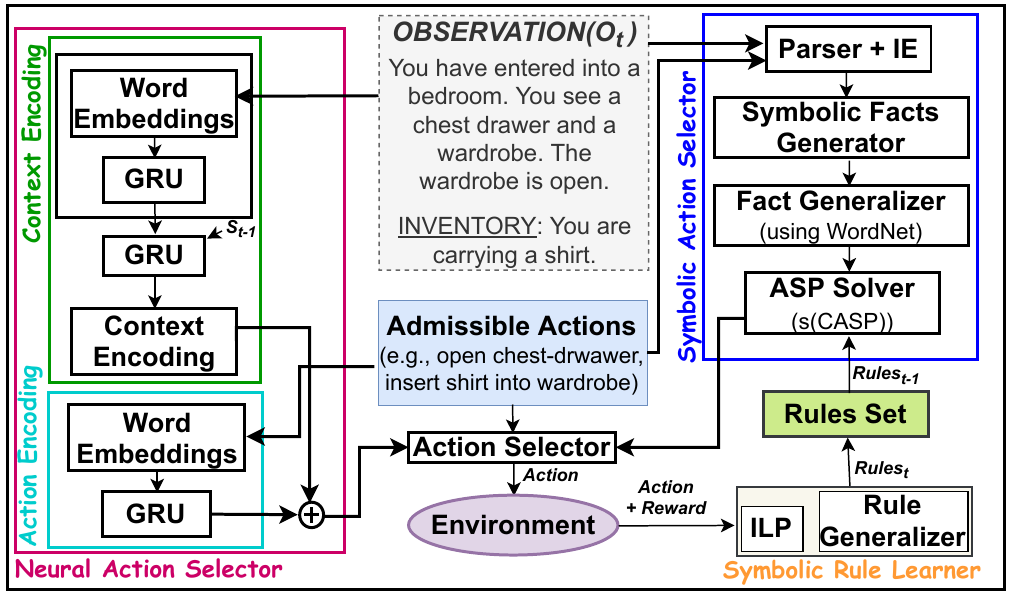}
    \caption{Overview of EXPLORER's decision-making at any given time step. The Hybrid Neuro-Symbolic architecture mainly consists of 5 modules - (a) \textcolor{Green}{Context Encoder} encodes the observation to dynamic context, (b) \textcolor{SkyBlue}{Action Encoder} encodes the admissible actions, (c) \textcolor{RedViolet!100}{Neural Action Selector} combines (a) and (b) with $\bigoplus$ operator, (d) \textcolor{blue}{Symbolic Action Selector} returns a set of candidate actions, and (e) \textcolor{BurntOrange}{Symbolic Rule Learner} uses ILP and WordNet-based rule generalization to generate symbolic rules.}
    \label{fig:hns_arch}
\end{figure}

Figure~\ref{fig:hns_arch} illustrates the components of our neuro-symbolic architecture and shows an overview of the agent's decision-making process. 

We have used TW-cooking to verify our approach and then performed a comprehensive evaluation of \ex on TWC games. To showcase the scalability aspects of EXPLORER, we have done comparative studies with other SOTA neural and neuro-symbolic models, and the empirical results demonstrate that \ex outplays others by achieving better generalization over unseen entities. Due to the neuro-symbolic nature of EXPLORER, we are also able to perform detailed qualitative studies of the policies (illustrated in section - \ref{sec:results}).

The main contributions of this paper are: (1) we present EXPLORER for TBGs that outperforms existing models in terms of steps and scores; (2) we discuss the importance of non-monotonic reasoning in  partially observable worlds; (3) we demonstrate how default theories can be learned with exceptions in an online manner for TBGs; and  (4) we provide a novel information-gain based rule generalization algorithm that leverages WordNet. 

\section{Background}
\label{sec:background}

\noindent\textbf{Text-based Reinforcement Learning:}
TBGs provide a challenging environment where an agent can observe the current state of the game and act in the world using only the modality of text. 
The agent perceives the state of the game only through natural language observations. Hence, TBGs can be modeled as a Partially Observable Markov Decision Process (POMDP) $(\mathcal{S}, \mathcal{A}, \mathcal{O}, \mathcal{T}, \mathcal{E}, r) $, where $\mathcal{S}$ is the set of states of the game, $\mathcal{A}$ is the natural language action space,
$\mathcal{O}$ is the set of textual observations describing the current state,
$\mathcal{T}$ are the conditional transition probabilities from one state to another, $\mathcal{E}$ are the conditional observation probabilities, 
$r: \mathcal{S} \times \mathcal{A} \rightarrow \mathbb{R}$ is a scalar reward function, which maps a state-action pair to the reward received by the agent.


\noindent\textbf{Inductive Logic Programming (ILP):}
 ILP is a machine learning technique where the learned model is in the form of logic programming rules (Horn Clauses) that are comprehensible to humans. It allows the background knowledge to be incrementally extended without requiring the entire model to be re-learned. Additionally, the comprehensibility of symbolic rules makes it easier for users to understand and verify induced models and even edit them. Details can be found in the work of~\citet{muggleton1994inductive}.


\noindent\textbf{Answer Set Programming (ASP): }
    An answer set program is a collection of rules of the form: 
 \[l_0 \leftarrow l_1, \;...\,, \;l_m, \; not \,  l_{m+1}, \;...\,, \; not\, l_n. \]
    
\noindent Classical logic denotes each  \( l_i\) is a literal \cite{gelfond2014knowledge}. 
ASP supports \textit{negation as failure} \cite{gelfond2014knowledge}, allowing it to elegantly model common sense reasoning, default rules with exceptions, etc. 


\noindent\textbf{s(CASP) Engine: }
For this work, we have used s(CASP) ASP solver to predict an action. s(CASP) \cite{scasp}  is a query-driven, goal-directed implementation of ASP that includes constraint solving over reals. Goal-directed execution of s(CASP) is indispensable for automating commonsense reasoning, as traditional grounding and SAT-solver based implementations of ASP may not be scalable. There are three major advantages of using the s(CASP) system: (i) s(CASP) does not ground the program, which makes our framework scalable, (ii) it only explores the parts of the knowledge base that are needed to answer a query, and (iii) it provides natural language justification  (proof tree) for an answer \cite{scasp_justification}.

\section{Symbolic Policy Learner} \label{sec:smb_pol_learn}





Deep reinforcement learning (DRL) has experienced great success by learning directly from high-dimensional sensory inputs, yet it suffers from a lack of interpretability. Interpretability of an agent's action is of utmost importance in sequential decision-making problems, as it increases the transparency of black-box-style agents; it also helps RL researchers understand the high-level behavior of the system better. To make a system interpretable, one of the most widely used approaches is learning the agent's policies symbolically. In our work, \ex learns these symbolic policies in the form of logical rules represented in the ASP. An example of such a rule is - \codec{insert(X, fridge) :- apple(X)} which can be translated as \textit{``X is insertable into a fridge if X is an apple''} \footnote{For ease of use, we retain action names as the predicate names; however, they are interpreted normally as adjectives.}. These learned ASP rules not only provide a better understanding of the system's functionality but can also be used to predict the agent's next action using an ASP solver. \ex learns the rules iteratively (in an online manner) and applies the rules to predict an action in collaboration with the neural module. Our results show that this approach is very effective in terms of performance and interpretability. 

\noindent\textbf{Partial Observability and Non-Monotonic Reasoning:}
\ex works in a partially observable environment, where it needs to predict an action based on its prior knowledge. If \ex fails, then it learns something new that will be applied in the next episode. The reasoning approach of \ex is non-monotonic in nature: that is, what it believes currently may become false in the future with new evidence.
We can model this using a non-monotonic logic programming paradigm that supports \textit{default rules} and \textit{exception to defaults} \cite{gelfond2014knowledge}. 
In this work, the belief of \ex has been represented as an Answer Set Program in the form of \textit{default rules} with \textit{exceptions}.  With the help of Inductive Logic Programming (ILP) (see Section~\ref{sec:ilp-smb-pol}) and Exception learner (see Section~\ref{sec:exc-learner}), these rules are learned by \ex after each episode and then applied in the following episode. The agent uses an ASP solver to predict actions by utilizing the observation and the rules. Based on the outcome after applying the rules, the learned policies are updated with the exception (if needed), and new rules are learned as needed.


\subsection{Learning Symbolic Policy using ILP}\label{sec:ilp-smb-pol} 

\begin{figure}[t]
    \centering
    \includegraphics[scale = 0.7]{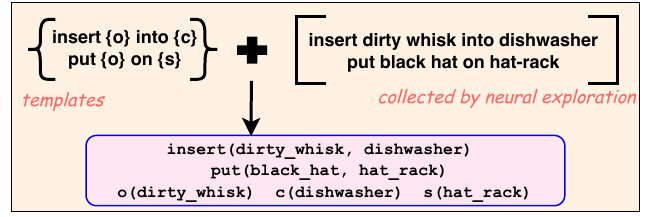}
    \caption{Entity extraction using Action Template}
    \label{fig:action_template}
\end{figure}

\noindent\textbf{{Data Collection:}}
To apply an ILP algorithm, first, \ex needs to collect the \textit{State}, \textit{Action}, and \textit{Reward} pairs while exploring the text-based environment. In a TBG, the two main components of the state are the state description and the inventory information of the agent. The entities present in the environment are extracted by parsing the state description using the {\tt spaCy} library, and only storing the noun phrases (e.g., {\tt fridge}, {\tt apple}, {\tt banana}, etc.) in predicate form. We also extract the inventory information in a similar way. At each step of the game, the game environment generates a set of admissible actions, one among them being the best; as well as action templates (e.g., \textit{``insert \code{O} into \code{S}''}, where \code{O} and \code{S} are entity types) which are predefined for the agent before the game starts. By processing these templates over the admissible actions, \ex can easily extract the type of each entity present in the environment and then convert them to predicates. Figure~\ref{fig:action_template} illustrates an instance of a predicate generation process. Along with this \textit{State} description, \ex also stores the taken \textit{Action} and the \textit{Reward} information at each step.

\noindent\textbf{{Data Preparation:}}
To learn the rules, an ILP algorithm requires three things - the \textit{goal}, the \textit{predicate list}, and the \textit{examples}. The \textit{goal} is the concept that the ILP algorithm is going to learn by exploring the examples. The \textit{predicates} give the explanation to a concept.  In the learned theory formulated as logical rules, \textit{goal} is the head and the \textit{predicate list} gives the domain space for the body clauses. The \textit{examples} are the set of positive and negative scenarios that are collected by the agent while playing.

\noindent\textbf{{Execution and Policy Learning:}}
In our work, we have mainly focused on learning the hypothesis for the rewarded actions; however, we also apply reward shaping to learn important preceding actions (e.g., \textit{open fridge} might not have any reward, although it is important to \textit{take an item from fridge} and that has a reward). In both the TW-Cooking domain and TextWorld Commonsense (TWC), the action predicates mostly have one or two arguments (e.g.,  \textit{open} \code{fridge}, \textit{insert} \code{cheese} \textit{in} \code{fridge}, etc.). 
In the one-argument setting, the \textit{action} becomes the ILP goal and the examples are collected based on the argument. In the two-argument setting, we fix the second argument with the action and collect examples based on the first argument. The goal will hence be in the form of \textit{\textless action\_(second\_argument) \textgreater}. We split the examples (i.e., state, entity types, inventory information in predicate form) based on the stored rewards (positive and zero/negative). We use entity identifiers to identify each entity separately; this is important when there are two or more instances of the same entity in the environment with different features (e.g., \textit{red apple} and \textit{rotten apple}). Additionally, \ex creates the predicate list by extracting the predicate names from the examples. After obtaining the goal, predicate list, and the example, the agent runs the ILP algorithm to learn the hypothesis, followed by simple string post-processing to obtain a hypothesis in the below form:

\smallskip
    
\cprotect 
        {
        \begin{minipage}{0.95\linewidth}
        { \code{action( X , entity) <- feature(X).}\\
          \code{action( X ) <- feature(X).}}
        \end{minipage}
}

\smallskip

\noindent 
Figure \ref{fig:ilp_example} elaborates the ILP data preparation procedure along with an example of a learned rule. 

\subsection{Exception Learning} \label{sec:exc-learner}


As \ex does online learning, the quality of the initial rules is quite low; this gradually improves with more training. The key improvement achieved by \ex is through exception learning, where an exception clause is added to the rule's body using Negation as Failure (NAF). This makes the rules more flexible and able to handle scenarios where information is missing. The agent learns these exceptions by trying the rules and not receiving rewards. For example, in TWC, the agent may learn the rule that - \textit{apple goes to the fridge}, but fail when it tries to apply the rule to a \textit{rotten apple}. It then learns that the feature {\it rotten} is an exception to the previously learned rule. This can be represented as:

\begin{figure}[t]
    \centering
    \includegraphics[scale = 0.65]{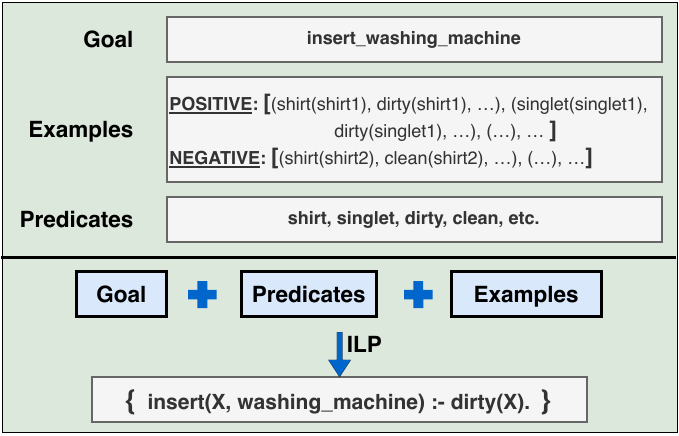}
    \caption{ILP Rule Learning Example}
    \label{fig:ilp_example}
\end{figure}

\smallskip
\noindent
\cprotect \fbox{
        \centering
        \begin{minipage}{0.95\linewidth}
        {\small \tt
insert(X , fridge) <- apple(X), not ab(X).
        ~~~ ab(X) <- rotten(X).}      
        \end{minipage}
    }
    
\smallskip
\noindent
It is important to keep in mind that the number of examples covered by the exception is always fewer than the number of examples covered by the defaults. This constraint has been included in EXPLORER's exception learning module.

\section{Rule Generalization}




\begin{figure}[t]
    \centering
    \includegraphics[scale = 0.58]{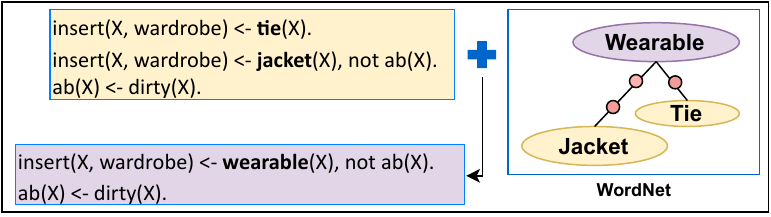}
    \caption{Example of Rule Generalization}
    \label{fig:gen_example}
\end{figure}

\smallskip
\noindent\textbf{{Importance of Rule Generalization:}}
An ideal RL agent should not only perform well on entities it has seen but also on unseen entities or out-of-distribution (OOD) data. To accomplish this, policy generalization is a crucial feature that an ideal RL agent should have. To verify this, we used \ex without generalization on the TW-Cooking domain, where it performs well, however, it struggles on the TWC games. TWC games are designed to test agents on OOD entities that were not seen during training but are similar to the training data. As a result, the policies learned as logic rules will not work on unseen objects.

For example, the rule for \textit{apple} (e.g., \codec{insert(X, fridge) \textless -  apple(X).}) cannot work on another fruit such as \textit{orange.} To tackle this, we lift the learned policies using WordNet's \cite{wn} hypernym-hyponym relations to get the generalized rules (illustration in  Figure \ref{fig:gen_example}). 
Motivation comes from the way humans perform tasks. For example, if we know a \textit{dirty shirt} goes to the \textit{washing machine} and we have seen a \textit{dirty pant}, we would put the \textit{dirty pant} into the \textit{washing machine} as both are of type \textit{clothes} and \textit{dirty}.

\begin{algorithm}[b]
    \small
    \caption{Generalized Rule Learner}\label{algorithm_wn}
    
     \hspace*{\algorithmicindent} \textbf{Input:} \textit{E}: \text{Examples (\textit{States, Actions, and Rewards)}}
    
    \hspace*{\algorithmicindent} \textbf{Output:} \textbf{$R_G$}: \text{Generalized Rules Set} 
    
    \begin{algorithmic}[1]
        \Procedure{GetGeneralizedRules}{E}
        \State $ \textit{$R_G$} ~ \gets ~ \textit{\{\}}$  \Comment{initialization}
        \State $ \textit{$Goals$} ~ \gets ~ \textit{getGoals(E)}$  ~~~~~~~~~~\Comment{get the list of goals similar to the ILP data preparation (described above)}
        \State \textbf{for each} {$g ~ \in ~ \textit{Goals}$} \textbf{do}
             \State \hspace{4mm} $ \textit{$E_g$} ~ \gets ~ \textit{getExamples(E, g)}$
             \State \hspace{4mm} $ \textit{$(E_g^+, E_g^-)$} ~ \gets ~ \textit{splitByRewards($E_g$)}$ 
             \State \hspace{4mm} $ \textit{$(Hyp_g^+, Hyp_g^-)$} ~ \gets ~ \textit{extractHypernyms($E_g^+, E_g^-$)}$  $\newline ~~~~~~~~~~$ \Comment{get the hypernyms from WordNet}
             \State \hspace{4mm} $ \textit{$r_g$} ~ \gets ~ \textit{getBestGen($E_g^+, E_g^-, Hyp_g^+, Hyp_g^-$)}$ $~~~~~~~~~$ \Comment{uses entropy based information gain formula}
             
             \State \hspace{4mm} $ \textit{$R_G$} ~ \gets ~ \textit{$R_G$}  \; \; \;  \cup \; \; \; \textit{$r_g$}$
             \State \textbf{end for}
        \State \Return{$R_G$}
        \EndProcedure
    \end{algorithmic}
\end{algorithm}

\smallskip
\noindent\textbf{{Excessive Generalization is Bad:}} On one hand, generalization results in better policies to work with unseen entities; however, too much generalization leads to a drastic increment in false-positive results. To keep the balance, \ex should know how much generalization is good. For an example, ``\textit{apple} is a \textit{fruit}'', ``\textit{fruits} are \textit{part of a plant}'', and ``\textit{plants} are \textit{living thing}''. Now, if we apply the same rule that explains a property of an \textit{apple} to all \textit{living things}, the generalization will have gone too far. So, to solve this, we have proposed a novel approach described in Section - \ref{sec:rule-gen-algo}.

\subsection{Dynamic Rule Generalization}\label{sec:rule-gen-algo}
In this paper, we introduce a novel algorithm to dynamically generate the generalized rules exploring the hypernym relations from WordNet (WN). The algorithm is based on information gain calculated using the entropy of the positive and negative set of examples (collected by EXPLORER). The illustration of the process is given in the Algorithm \ref{algorithm_wn}. The algorithm takes the collected set of examples and returns the generalized rules set. First, similar to the ILP data preparation procedure, the \textit{goals} are extracted from the examples. For each \textit{goal}, examples are split into two sets - $E^+$ and $E^-$. Next, the hypernyms are extracted using the hypernym-hyponym relations of the WordNet ontology. The combined set of hypernyms from ($E^+$, $E^-$) gives the \textit{body} predicates for the generalized rules. Similar to the ILP (discussed above) the \textit{goal} will be the \textit{head} of a generalized rule. Next, the best-generalized rules are generated by calculating the max information gain between the hypernyms. Information gain for a given clause is calculated using the below formula \cite{mitchell} ---

\begin{equation*}
IG(R, h) = total * ( log_2 \frac{p_1}{p_1 + n_1} -  log_2 \frac{p_0}{p_0 + n_0} )
\end{equation*}

\noindent
where $h$ is the candidate hypernym predicate to add to the rule $R$, $p_0$ is the number of positive examples implied by the rule $R$, $n_0$ is the number of negative examples implied by the rule $R$, $p_1$ is the number of positive examples implied by the rule $R + h$, $n_1$ is the number of negative examples implied by the rule $R + h$, $total$ is the number of positive examples implied by $R$ also covered by $R + h$. Finally, it collects all the generalized rules set and returns. It is important to mention that this algorithm only learns the generalized rules which are used in addition to the rules learned by ILP and exception learning (discussed in section \ref{sec:smb_pol_learn}).

\section{Experiments and Results}
\label{sec:experiments}

\subsection{Dataset}
In our work, we want to show that if an RL agent uses symbolic and neural reasoning in tandem, where the neural module is mainly responsible for exploration and the symbolic component for exploitation, then the performance of that agent increases drastically in text-based games. At first, we verify our approach with TW-Cooking domain \cite{tw-cook}, where we have used levels 1-4 from the GATA dataset\footnote{https://github.com/xingdi-eric-yuan/GATA-public} for testing. As the name suggests, this game suit is about collecting  various cooking ingredients and preparing a meal following an in-game recipe. 


To showcase the importance of generalization, we have tested our \ex agent on TWC games with OOD data. Here, the goal is to tidy up the house by putting objects in their commonsense locations. With the help of TWC framework \cite{twc_aaai}, we have generated a set of games with 3 different difficulty levels - (i) \textit{easy level:} that contains 1 room with 1 to 3 objects; (ii)\textit{ medium level:} that contains 1 or 2 rooms with 4 or 5 objects; and (iii) \textit{hard level:} a mix of games with a high number of objects (6 or 7 objects in 1 or 2 rooms) or a high number of rooms (3 or 4 rooms containing 4 or 5 objects).

We chose TW-Cooking and TWC games as our test-bed because these are benchmark datasets for evaluating neuro-symbolic agents in text-based games \cite{chaudhury2021neuro, chaudhury2023learning, wang2022behavior, kimura2021neuro, basu2022hybrid}. Also, these environments require the agents to exhibit skills such as exploration, planning, reasoning, and OOD generalization, which makes them ideal environments to evaluate EXPLORER.



\begin{table*}[t]
\centering
\scriptsize
\begin{tabular}{|c|c|c|c|c|c|c|c|c|}
\hline
\multicolumn{3}{|c|}{\multirow{2}{*}{}}                                                                                & \multicolumn{2}{c|}{Easy}    & \multicolumn{2}{c|}{Medium}  & \multicolumn{2}{c|}{Hard}    \\ \cline{4-9} 
\multicolumn{3}{|c|}{}                                                                                                 & \#steps       & N. Score     & \#steps       & N. Score     & \#steps       & N. Score     \\ \hline
\multirow{6}{*}{IN}  & \multicolumn{2}{c|}{Text Only}                                                                  & \cellcolor{LimeGreen!30} 15.12 $\pm$ 1.95 &  \cellcolor{LimeGreen!80} 0.91 $\pm$ 0.03 & \cellcolor{Yellow!40} 33.17 $\pm$ 2.76 & \cellcolor{LimeGreen!50} 0.83 $\pm$ 0.04 & \cellcolor{Red!50} 47.68 $\pm$ 2.43 & \cellcolor{YellowOrange!40} 0.6 $\pm$ 0.05  \\ \cline{2-9} 
                     & \multicolumn{2}{c|}{EXPLORER-w/o-GEN}                                                           & \cellcolor{LimeGreen!30} 17.39 $\pm$ 3.01 &  \cellcolor{LimeGreen!80} 0.93 $\pm$ 0.04 & \cellcolor{Red!50} 46.7 $\pm$ 2.14  & \cellcolor{Red!40} 0.42 $\pm$ 0.12 &  \cellcolor{Yellow!60}  37.66 $\pm$ 0.93 & \cellcolor{LimeGreen!50} 0.88 $\pm$ 0.01 \\ \cline{2-9} 
                     & \multirow{3}{*}{\begin{tabular}[c]{@{}c@{}} \ex \end{tabular}} & Exhaustive       & \cellcolor{LimeGreen!50} 12.86 $\pm$ 3.04 & \cellcolor{LimeGreen!80} 0.91 $\pm$ 0.04 & \cellcolor{Yellow!20} 29.9 $\pm$ 3.16  & \cellcolor{YellowOrange!40} 0.65 $\pm$ 0.06 & \cellcolor{Yellow!40} 30.44 $\pm$ 0.87 & \cellcolor{LimeGreen!80} 0.95 $\pm$ 0.03 \\ \cline{3-9} 
                     &                                                                              & IG (Hyp. Lvl. 2) & \cellcolor{LimeGreen!50} 10.59 $\pm$ 1.3  & \cellcolor{LimeGreen!80} 0.95 $\pm$ 0.02 & \cellcolor{YellowGreen!20}22.57 $\pm$ 1.04 &  \cellcolor{GreenYellow!80} 0.77 $\pm$ 0.07 & \cellcolor{Yellow!40} 30.46 $\pm$ 0.74 & \cellcolor{LimeGreen!50} 0.87 $\pm$ 0.01   \\ \cline{3-9} 
                     &                                                                              & IG (Hyp. Lvl. 3) & \cellcolor{LimeGreen!70} 9.55 $\pm$ 2.34  &  \cellcolor{LimeGreen!80} 0.96 $\pm$ 0.02 & \cellcolor{Yellow!20} 25.34 $\pm$ 2.86 & \cellcolor{GreenYellow!80} 0.76 $\pm$ 0.03 & \cellcolor{Yellow!40} 33.54 $\pm$ 1.47 & \cellcolor{LimeGreen!80} 0.91 $\pm$ 0.03 \\ \hline \hline
\multirow{6}{*}{OUT} & \multicolumn{2}{c|}{Text Only}                                                                  & \cellcolor{LimeGreen!30} 16.66 $\pm$ 1.74 & \cellcolor{LimeGreen!80} 0.92 $\pm$ 0.03 & \cellcolor{Yellow!60} 37.3 $\pm$ 3.45  & \cellcolor{GreenYellow!80} 0.73 $\pm$ 0.06 & \cellcolor{Red!70} 50.00 $\pm$ 0.0  & \cellcolor{Red!50} 0.3 $\pm$ 0.04  \\ \cline{2-9} 
                     & \multicolumn{2}{c|}{EXPLORER-w/o-GEN}                                                           & \cellcolor{YellowGreen!20} 21.19 $\pm$ 0.87 & \cellcolor{LimeGreen!50} 0.84 $\pm$ 0.06 & \cellcolor{Red!50} 46.36 $\pm$ 1.52 & \cellcolor{Red!40} 0.42 $\pm$ 0.08 & \cellcolor{YellowOrange!50}44.25 $\pm$ 0.42 & \cellcolor{YellowOrange!40} 0.63 $\pm$ 0.01 \\ \cline{2-9} 
                     & \multirow{3}{*}{\begin{tabular}[c]{@{}c@{}}EXPLORER\end{tabular}} & Exhaustive       & \cellcolor{LimeGreen!50} 14.65 $\pm$ 2.18 & \cellcolor{LimeGreen!80} 0.91 $\pm$ 0.05 &  \cellcolor{Yellow!60}  37.07 $\pm$ 2.09 & \cellcolor{YellowOrange!40} 0.63 $\pm$ 0.06 & \cellcolor{YellowOrange!50} 41.52 $\pm$ 1.12 & \cellcolor{LimeGreen!50} 0.83 $\pm$ 0.02 \\ \cline{3-9} 
                     &                                                                              & IG (Hyp. Lvl. 2) & \cellcolor{LimeGreen!30} 15.08 $\pm$ 1.2  & \cellcolor{LimeGreen!80} 0.91 $\pm$ 0.02 &  \cellcolor{YellowOrange!50} 40.63 $\pm$ 3.03 & \cellcolor{Orange!50} 0.57 $\pm$ 0.06 & \cellcolor{YellowOrange!50} 42.18 $\pm$ 0.66 & \cellcolor{GreenYellow!80} 0.79 $\pm$ 0.01             \\ \cline{3-9} 
                     &                                                                              & IG (Hyp. Lvl. 3) & \cellcolor{LimeGreen!50} 12.72 $\pm$ 1.22 & \cellcolor{LimeGreen!80} 0.92 $\pm$ 0.02 &  \cellcolor{Yellow!60}  37.38 $\pm$ 3.09 & \cellcolor{YellowOrange!40} 0.64 $\pm$    0.09 & \cellcolor{YellowOrange!50}43.16 $\pm$ 2.83 & \cellcolor{GreenYellow!80} 0.78 $\pm$ 0.03 \\ \hline
    \end{tabular}
    \caption{ TWC performance comparison results for within distribution (IN) and out-of-distribution (OUT) games}
    \label{tab:results_twc}
\end{table*}

\subsection{Experiments}
To explain \ex works better than a neural-only agent, we have selected two neural baseline models for each of our datasets (TWC and TW-Cooking) and compared them with \textbf{EXPLORER}. In our evaluation, for both the datasets, we have used  LSTM-A2C \cite{narasimhan2015language} as the \textbf{Text-Only} agent, which uses the encoded history of observation to select the best action. For TW-Cooking, we have compared \ex with the SOTA model on the TW-Cooking domain - Graph Aided Transformer Agent (\textbf{GATA}) \cite{tw-cook}. Also, we have done a comparative study of neuro-symbolic models on TWC (section \ref{sec:results}) with SOTA neuro-symbolic model \textbf{CBR} \cite{cbr}, where we have used SOTA neural model \textbf{BiKE} \cite{bike} as the neural module in both EXPLORER and CBR.

We have tested with four neuro-symbolic settings of EXPLORER, where one without generalization - \textbf{EXPLORER-w/o-GEN} and the other three uses \ex with different settings of generalization. Below are the details of different generalization settings in EXPLORER:

\noindent \textbf{{Exhaustive Rule Generalization:}} This setting lifts the rules exhaustively with all the hypernyms up to WordNet level 3 from an object or in other words select those hypernyms of an object whose path-distance with the object is $\leq$ 3.    

\noindent \textbf{{IG-based generalization (hypernym Level 2/3):}} Here, \ex uses the rule generalization  algorithm (algorithm \ref{algorithm_wn}). It takes WordNet hypernyms up to level 2 or 3 from an object.


For both datasets in all the settings, agents are trained using 100 episodes with 50 steps maximum. On TW-Cooking domain, it is worth mentioning that while we have done the pre-training tasks (such as graph encoder, graph updater, action scorer, etc) for GATA as in \cite{tw-cook}, both text-only agent and EXPLORER do not have any pre-training advantage to boost the performance.

\begin{table*}[t]
    \centering
    \small
    \begin{tabular}{|c|c|c|c|c|c|c|}
    \hline
    \multicolumn{1}{|l|}{\multirow{2}{*}{}} & \multicolumn{2}{c|}{\begin{tabular}[c]{@{}c@{}}Text-Only \\ (Neural)\end{tabular}} & \multicolumn{2}{c|}{\begin{tabular}[c]{@{}c@{}} GATA \\ (Neural)\end{tabular}} & \multicolumn{2}{c|}{\begin{tabular}[c]{@{}c@{}} EXPLORER-w/o-GEN \\ (Neuro-Symbolic) \end{tabular}} \\ \cline{2-7} 
    \multicolumn{1}{|c|}{}                  & \multicolumn{1}{c|}{\#Steps}            & \multicolumn{1}{c|}{N. Score}      & \multicolumn{1}{c|}{\#Steps}            & \multicolumn{1}{c|}{N. Score}     & \multicolumn{1}{c|}{\#Steps}          & \multicolumn{1}{c|}{N. Score}          \\ \hline
    Level-1 & 11.50 $\pm$ 1.26 & 0.93 $\pm$ 0.01 & 12.01 $\pm$ 0.84 & 0.72 $\pm$ 0.3 & \textbf{9.17 $\pm$ 3.28} & \textbf{0.96 $\pm$ 0.04} \\ 
    Level-2 & 45.79 $\pm$ 1.56 & 0.35 $\pm$ 0.07 & 28.65 $\pm$ 1.28 & 0.23 $\pm$ 0.06 & \textbf{15.60 $\pm$ 3.74} & \textbf{0.58 $\pm$ 0.04} \\ 
    Level-3 & 46.91 $\pm$ 1.51 & 0.25 $\pm$ 0.01 & 37.6 $\pm$ 1.17 & 0.18 $\pm$ 0.07 & \textbf{26.92 $\pm$ 2.74} & \textbf{0.34 $\pm$ 0.01} \\ 
    Level-4 & 22.03 $\pm$ 0.19 & 0.76 $\pm$ 0.01 & 35.53 $\pm$ 2.5 & 0.34 $\pm$ 0.06 & \textbf{19.85 $\pm$ 2.12} & \textbf{0.85 $\pm$ 0.07}\\ \hline

    \end{tabular}
    \caption{ TW-Cooking domain --- Comparison Results (with Mean and SD)}
    \label{tab:results_cooking_full}
\end{table*}

\subsection{Results}\label{sec:results}
In our experiments, we evaluated the models based on the number of steps taken by the agent - \code{\#steps} (lower is better) and the normalized scores - \code{n. score} (higher is better). 
For TWC, Table \ref{tab:results_twc} shows the comparison results of all 4 settings along with the baseline model (Text-Only agent). We compared our agents in two different test sets - (i) \textit{IN distribution:} that has the same entities (i.e., objects and locations) as the training dataset, and (ii) \textit{OUT distribution:} that has new entities, which have not been included in the training set.
Table \ref{tab:results_cooking_full} illustrates the results on TW-Cooking.  The purpose of TW-Cooking is only to verify our approach and for that, we have used EXPLORER-w/o-GEN as the neuro-symbolic setting. 

For each result (shown in Tables \ref{tab:results_twc} and \ref{tab:results_cooking_full}), the \code{\#steps} and \code{n. score} should be seen together to decide which agent is doing better than the others. On one hand, if we focus more on the \code{\#steps}, the agents can be given very low max steps to complete a game, where the agents will perform well in terms of \code{\#steps} (lower is better), however the \code{n. score} will be very low (higher is better) as most of the games are not completed. On the other hand, if we focus more on \code{n. score}, the agent can be given very high max steps to complete a game, where the agent will score very high \code{n. score}. As both the cases are wrong interpretations of results, we should consider taking into account both \code{\#step} and \code{n. score} to judge an agent’s performance.

\noindent {{\textbf{{Qualitative Studies:}}}}
In our verification dataset - TW-cooking domain games, we found that the \ex does really well and beats the Text-Only and GATA agents in terms of \#step and normalized scores on all the levels. In level-1 which is focused on only collecting the ingredients, \ex does slightly better than the Text-Only agents as the neural models are already good in easier games (with less sequence of actions) so the scope of improvement is less for the EXPLORER. However, as the level increases in level-2 and 3, where it requires collecting ingredients, processing them and cooking to prepare the meal, and finally eating them to complete the game, the importance of neural-based exploration and symbolic-based exploitation comes into play. In Level-4, which includes navigation, the neural module helps the \ex to navigate to the kitchen (i.e., \textit{exploration}) from one part of the house, and then the symbolic module applies its rules to choose the best action (i.e., \textit{exploitation}).

In TWC, \ex with \textit{IG based generalization (hypernym Level 3)} performs better than the others in the \textit{easy} and \textit{medium} level games, whereas \textit{exhaustive generalization} works well in the \textit{hard} games. This shows that on one side, the exhaustive generalization works slightly better in an environment where the entities and rooms are more, and that needs more exploration. On another side, IG-based generalization works efficiently when the agent's main task is to select appropriate locations of different objects. In the \textit{easy} and \textit{medium} games, the EXPLORER-wo-GEN performs poorly in comparison with the baseline model. This indicates - only learning rules without generalization for simple environments leads to bad action selection especially when the entities are unseen. The out-distribution results for the medium games are not up to the mark. Further studies on this show that this happens when the OOD games have different but similar locations (\textit{clothes-line} vs. \textit{clothes-drier}) along with different objects in the environment. Generalization on the location gives very noisy results (increases false-positive cases) as they already belong to a higher level in the WordNet ontology. One of the solutions to this problem is to use a better neural model for exploration which helps to learn better rules (shown in the \textit{Comparative Study} sub-section in Section - \ref{sec:results}). Another solution for this issue requires a different way of incorporating commonsense to the agent and we have addressed more on this in the future work section.

\begin{figure}[t]
    \centering
    \includegraphics[scale = 1.0]{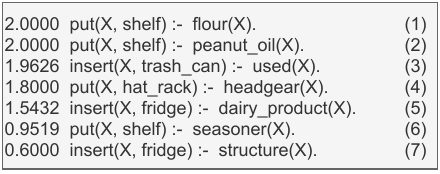}
    \caption{Example of Rule's Confidence Scores (medium level games) }
    \label{fig:gen_rule_conf}
\end{figure}

In the process of learning rules, we found that the symbolic agent is having difficulties choosing between multiple recommended (by the ASP solver) symbolic actions. So, it becomes an utmost importance to have a confidence score for each rule and we have generated that by calculating the accuracy. The accuracy of a rule can be calculated by the number of times the rule gives a positive reward divided by the total number of times the rule has been used. For generalized rules we have added another component for confidence calculation - that is how close the words are in WordNet, that is the score is reversely proportional to the distance between two nodes (entity and the hypernym). So for the non-generalized rules, the max score is 1, whereas for the generalized rules, the max score is 2 (due to the additional component). Figure \ref{fig:gen_rule_conf} shows a snippet of a learned set of generalized rules with different confidence scores.



\begin{figure}[t]
    \centering
    \includegraphics[scale = 0.86]{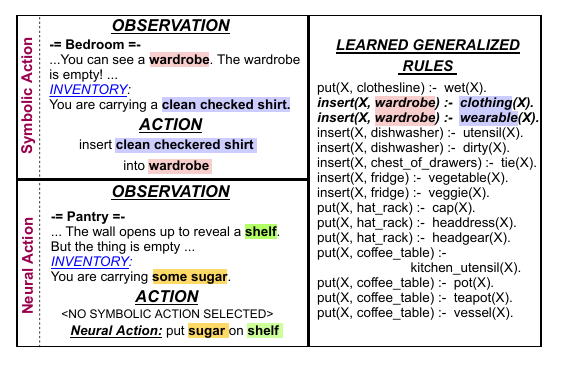}
    \caption{Examples from TWC game, showing the learned rules (right-hand side) along with the \textit{observations} and \textit{action selection} (Symbolic vs. Neural) }
    \label{fig:game_example}
\end{figure}

\noindent {{\textbf{{Example:}}}}
Figure \ref{fig:game_example} illustrates an example showing how the \ex plays the TWC games. On the right-hand side of the diagram, a snippet of the symbolic rules is given that the agent learns using ILP and ILP + WordNet-based rule generalization for TWC. 
To generate action using the symbolic module, the agent first extracts the information about the entities from the observation and the inventory information. Then, this information is represented as ASP facts along with the hypernyms of the objects. Next, it runs the ASP solver - the s(CASP) engine to get a set of possible actions and select an action based on the confidence scores. The top-left section of the figure \ref{fig:game_example} shows how a symbolic action has been selected by matching the object (i.e., \textit{`clean checked shirt'}) with the rule set (highlighted on the right). Here, the solver finds the location of the \textit{`shirt'} in \textit{`wardrobe'} as \textit{`clothing'} and \textit{`wearable'} are the hypernyms of the word - \textit{`shirt'}. In EXPLORER, when the symbolic agent fails to provide a good action, the neural agent serves as a fallback. In other words, \ex accords priority to the symbolic actions based on the confidence scores. \ex fallbacks to the neural actions when either of the one following situations happens: (i) no policies for the given state because of early exploration stage or non-rewarded actions, or (ii) non-admissible symbolic action has been generated due to rule generalization. The bottom-left of the figure \ref{fig:game_example} shows that the location of \textit{`sugar'} is not covered by any rules, so the neural agent selects an action. \ex learns rules in an online manner after each episode, so after the current episode, \ex will add the rule for sugar and in the next episode it will become a symbolic action. In this way, both the neural and symbolic modules work in tandem, where the neural module facilitates improved \textit{exploration} and the symbolic module helps to do better \textit{exploitation} in EXPLORER.

\begin{figure*}[t]
  \centering
  \begin{subfigure}{0.5\columnwidth}
    \centering
    \includegraphics[width=\linewidth]{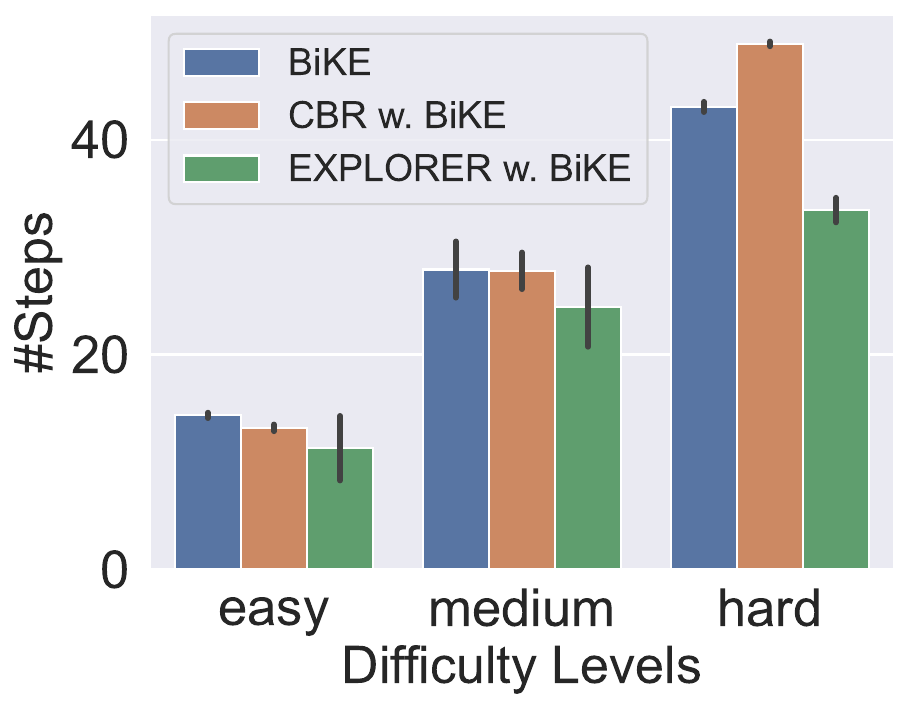}
    \caption{In Dist.: \#Steps}
  \end{subfigure}%
  \hfill
  \begin{subfigure}{0.5\columnwidth}
    \centering
    \includegraphics[width=\linewidth]{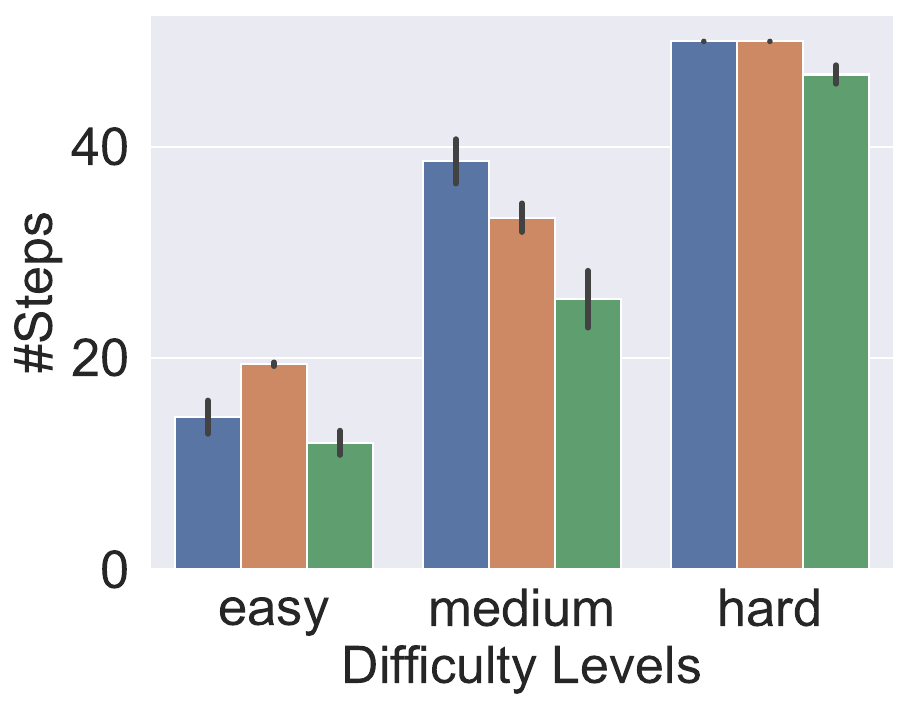}
    \caption{Out Dist.: \#Steps}
  \end{subfigure}%
  \hfill
  \begin{subfigure}{0.5\columnwidth}
    \centering
    \includegraphics[width=\linewidth]{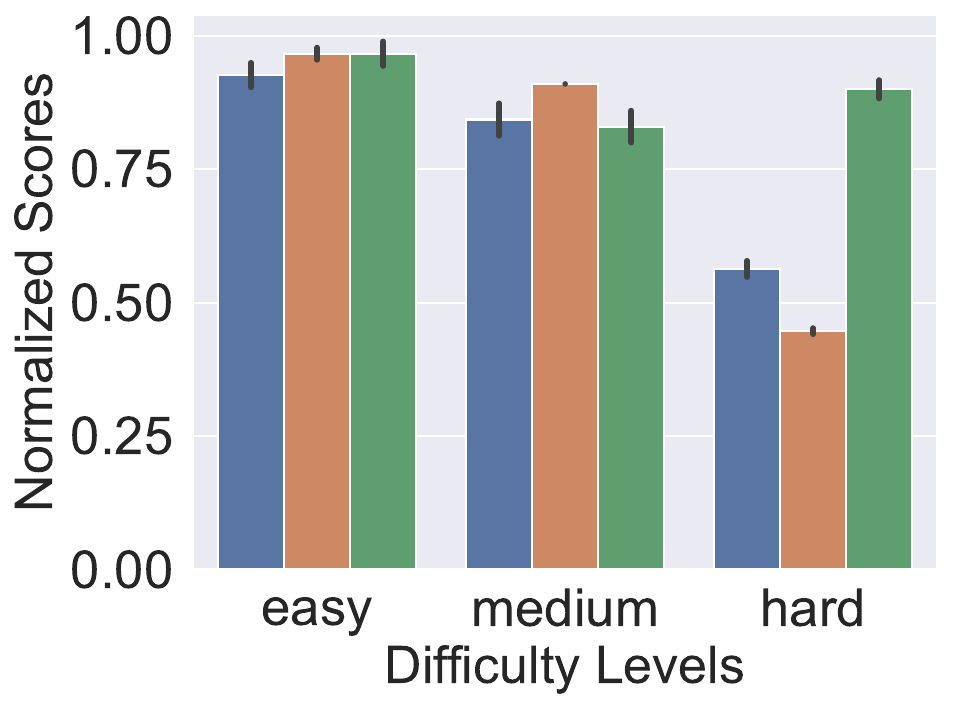}
    \caption{In Dist.: N. Score}
  \end{subfigure}
  \begin{subfigure}{0.5\columnwidth}
    \centering
    \includegraphics[width=\linewidth]{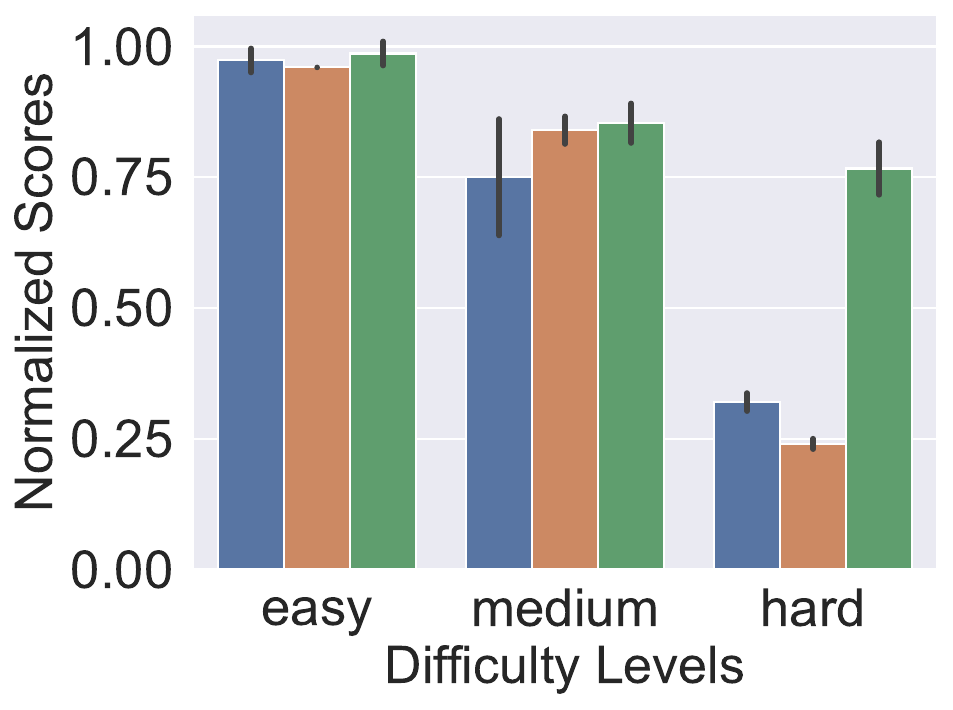}
    \caption{Out Dist.: N. Score}
  \end{subfigure}
  \caption{Performance on TWC: \textbf{BiKE}, \textbf{CBR w. BiKE}, and \textbf{EXPLORER w. BiKE}. Plot (a) and (b) show \#steps comparison of in and out distribution data (\textit{lower is better}); and plot (c) and (d) show the normalized scores comparison numbers of in and out distribution data (\textit{higher is better})}
\label{fig:bar-bike}
\end{figure*}

\noindent {{\textbf{{Comparative Studies:}}}}
One of the key contributions of \ex is that it is scalable and we can use any neural model as its base. To demonstrate that we have taken \textbf{BiKE} \cite{bike} a neural model designed for textual reinforcement learning. Then, we train EXPLORER (with Rule Generalization (hypernym level 3)) using BiKE as its neural module (instead of LSTM-A2C) and build \textbf{EXPLORER w. BiKE} agent. In this comparison study, we have also taken the neuro-symbolic SOTA baseline on TWC - Case-Based Reasoning (CBR) \cite{cbr} model and trained it with BiKE as its neural module as well and crafted \textbf{CBR w. BiKE} agent. Now we tested these 3 models over TWC games including \textit{easy}, \textit{medium}, and \textit{hard} levels. The performance evaluations are showcased with bar-plots in figure \ref{fig:bar-bike}. It clearly shows EXPLORER w. BiKE is doing much better at all levels in terms of \#Steps (\textit{lower is better}) and normalized scores (\textit{higher is better}). Also, EXPLORER w. BiKE is outperforming others with a large margin in out-distribution cases. This clearly depicts the importance of the policy generalization, which is helping the EXPLORER w. BiKE agent to use commonsense knowledge to reason over unknown entities. In the easy-level games, the performance differences are not that huge, as the environment deals with only one to three objects in a single room, which becomes much easier for the neural agent. However, as the level increases, we can start clearly seeing the importance of the EXPLORER agent.


\section{Related Work}
\label{sec:related}

\noindent\textbf{Text-based Reinforcement Learning:}
TBGs have recently emerged as promising environments for studying grounded language understanding and have drawn significant research interest.
\citet{zahavy2018learn} introduced the Action-Elimination Deep Q-Network (AE-DQN), which learns to predict invalid actions in the text-adventure game \textit{Zork}. 
\citet{cote18textworld} designed \texttt{TextWorld}, a sandbox learning environment for training and evaluating RL agents on text-based games. Building on this, \citet{twc_aaai} introduced \textit{TWC}, a set of games requiring agents with commonsense knowledge.
The \textit{LeDeepChef} system~\citep{Adolphs2019LeDeepChefDR} 
achieved good results on the \emph{First TextWorld Problems} \citep{ftwp} by supervising the model with entities from {\tt FreeBase}, allowing the agent to generalize to unseen objects.
A recent line of work learns symbolic (typically graph-structured) representations of the agent's belief. Notably, \citet{ammanabrolu2019playing} proposed \textit{KG-DQN} and \citet{adhikari2020learning} proposed \textit{GATA}. 
The \textit{following instruction for TBGs}  paper \cite{ltl_gata}, which was also focused on the TW-Cooking domain, assumes a lot about the game environment and provides many manual instructions to the agent. In our work, \ex automatically learns the rules in an online manner. 

\noindent\textbf{Symbolic Rule Learning Approaches:}
Learning symbolic rules using inductive logic programming has a long history of research. After the success of ASP, many works have emerged that are capable of learning non-monotonic logic programs, such as FOLD \cite{fold}, ILASP \cite{ilasp}, XHAIL \cite{xhail}, ASPAL \cite{aspal}, etc. However, there are not many efforts that have been taken to lift the rules to their generalized version and then learn exceptions. Also, they do not perform well on noisy data. To tackle this issue, there are efforts to combine ILP with differentiable programming \cite{dilp, end_dilp}. However, it requires lots of data to be trained on. In our work, we use a simple information gain based inductive learning approach, as the \ex learns the rules after each episode with a very small amount of examples (sometimes with zero negative examples).

\section{Future Work and Conclusion}
In this paper, we propose a neuro-symbolic agent EXPLORER that demonstrates how symbolic and neural modules can collaborate in a text-based RL environment. Also, we present a novel information gain-based rule generalization algorithm. Our approach not only achieves promising results in the TW-Cooking and TWC games but also generates interpretable and transferable policies. Our current research has shown that excessive reliance on the symbolic module and heavy generalization may not always be beneficial, so our next objective is to develop an optimal strategy for switching between the neural and symbolic modules to enhance performance.


\section*{Limitations}
One limitation of \ex model is its computation time, which is longer than that of a neural agent. \ex takes more time because it uses an ASP solver and symbolic rules, which involve multiple file processing tasks. However, the neuro-symbolic agent converges faster during training, which reduces the total number of steps needed, thereby decreasing the computation time difference between the neural and neuro-symbolic agents.

\section*{Ethics Statement}
In this paper, we propose a neuro-symbolic approach for text-based games that generates interpretable symbolic policies, allowing for transparent analysis of the model's outputs. Unlike deep neural models, which can exhibit language biases and generate harmful content such as hate speech or racial biases, neuro-symbolic approaches like ours are more effective at identifying and mitigating unethical outputs. The outputs of our model are limited to a list of permissible actions based on a peer-reviewed and publicly available dataset, and we use WordNet, a widely recognized and officially maintained knowledge base for NLP, as our external knowledge source. As a result, the ethical risks associated with our approach are low.

\bibliography{anthology,custom}

\end{document}